\documentclass{iopjournal}

\usepackage{booktabs}
\usepackage{algorithm}
\usepackage{algpseudocode}
\usepackage{float}
\usepackage{amsmath}

\begin{document}

\articletype{Paper} 

\title{Embedding Consistent Recurrent Neural Networks in Finite Element Simulations for Path-Dependent Damage Prediction}

\author{Muhammed Adil Yatk{\i}n$^{1,*}$\orcid{0000-0001-6776-8813}, Mihkel K{\~o}rgesaar$^1$\orcid{0000-0003-3920-9031} and Hasan Kurban$^2$\orcid{0000-0003-3142-2866}}

\affil{$^1$School of Engineering, Kuressaare College, Tallinn University of Technology, Tallinna 19, Kuressaare, 93811, Saaremaa, Estonia}
\affil{$^2$College of Science and Engineering, Hamad Bin Khalifa University, Education City, P.O. Box 34110, Doha, Qatar}
\affil{$^*$Author to whom any correspondence should be addressed.}

\email{muyatk@taltech.ee}


\begin{abstract}
Capturing sheet-metal localisation physics with a trained surrogate inside a finite element (FEM) solver requires the surrogate to deliver consistent predictions irrespective of the solver's strain-increment count, which is difficult to control in non-linear explicit codes. We address this by re-implementing a trained recurrent neural network (RNN) damage criterion in Fortran and embed it as a live, increment-by-increment fracture criterion inside an Abaqus/Explicit user material subroutine (VUMAT), advancing the network state alongside the solver. Two architectures are compared: a SimpleRNN and the proposed Consistent RNN (ConsRNN), whose transition function is designed so that predictions converge as the strain path is refined rather than drifting with the increment count. Both are trained on bilinear strain paths and evaluated under varying temporal discretisations and nonlinear histories. SimpleRNN predictions drift as the increment count increases, which disqualifies it for embedding; ConsRNN converges under refinement, at a substantial fixed-resolution accuracy cost relative to SimpleRNN on multilinear paths. Deployed at structural scale, the embedded ConsRNN surrogate governs element deletion in a clamped steel plate and reaches the same peak force and peak displacement as an established two-parameter fracture criterion. This pre-peak agreement is expected by construction, since no element is deleted before the peak and the two runs therefore share an identical constitutive path; the responses diverge only on the post-peak softening branch. The results establish discretisation consistency as the decisive property for embedding recurrent surrogates in explicit FEM solvers, and report the first embedding of a customised, non-standard-transition recurrent cell at structural scale.
\end{abstract}

\section{Introduction}
\label{intro}

Machine learning models are increasingly used as surrogates for the nonlinear relationship between deformation and mechanical response in engineering structures \cite{bonatti2022cp, khandelwal2021machine, suttakul2024role}. Instead of solving explicitly defined constitutive relations within finite element (FE) frameworks, data-driven models learn the input--output mapping directly from simulation data \cite{khandelwal2021machine}, motivating hybrid frameworks in which individual simulation components \cite{mozaffar2019deep, wu2020recurrent, ghavamian2019accelerating, MAIA2023115934, BONATTI2022104697}, or even entire simulation stages \cite{flaschel2021unsupervised, carrara2021data, kirchdoerfer2016data, eggersmann2019model}, are replaced by learned surrogates. Such models enable expensive simulation knowledge to be reused efficiently across analyses, while a single network formulation can represent multiple material classes with transferable learned representations \cite{doi:10.1126/sciadv.abf3658}. In this spirit, Yatk\i n and K\~{o}rgesaar \cite{YATKIN_P1} introduced a mapping from the deformation history of a material unit, sized to a large shell element, to its strain localisation response and demonstrated that this relationship can be learned directly from high-fidelity simulation data. In the present work, we take this localisation surrogate from offline validation to deployment, embedding a consistency-preserving recurrent network directly inside an explicit finite element solver.

Recurrent neural networks (RNNs) are natural candidates for learning such history-dependent mappings because their internal state advances across loading increments in the same way an FE solver advances its solution. Unlike fully connected neural networks, which require deformation history to be represented through hand-crafted state variables \cite{palau2012neural, gorji2019towards}, recurrent architectures naturally retain historical information within their hidden state and have successfully learned homogenised constitutive behaviour \cite{mozaffar2019deep}, arbitrary non-proportional loading paths \cite{GORJI2020103972}, and parametrized finite element simulations with varying mesh topologies \cite{GREVE2022100137}. However, predictive accuracy alone is insufficient for deployment as a finite element surrogate. Since explicit FE solvers determine their own increment sizes, surrogate predictions must remain consistent under temporal refinement. This has motivated the development of specialized recurrent transition functions capable of producing consistent estimates during simulation. Bonatti and Mohr \cite{BONATTI2022104697} established self-consistency as a fundamental property of recurrent transition functions intended for use as surrogates inside FEM-based simulations, and Yatkin et al. \cite{yatkin2026self} showed that any sequential learning architecture deployed as a surrogate must satisfy the truncation and consistency requirements.

This observation motivated several consistency-preserving recurrent formulations, including the Linearized Minimal State Cell \cite{BONATTI2022104697}, its rate- and temperature-dependent extension \cite{HEIDENREICH2025104305}, recurrent surrogates for coupled plasticity and fracture in composites \cite{GIRARD2026113703}, and self-consistent gated recurrent units with fewer parameters \cite{wu2024self}. Building upon this direction, our earlier studies \cite{Yatkin_ICMLA2023, Yatkin2025_NN_conference, korgesaar2025machine} progressively extended the localisation surrogate to recurrent architectures, culminating in the Consistent Recurrent Neural Network (ConsRNN) \cite{app15020933}. ConsRNN differs from the Linearized Minimal State Cell in construction. The LMSC achieves self-consistency by learning a compact internal state whose update rule embeds physics-motivated variables tailored to the constitutive response. ConsRNN, in contrast, retains a standard recurrent hidden state and secures consistency purely through its mathematical design: a norm-scaled convex-combination state-transition update which can be interpreted as an explicit-Euler discretisation of an arc-length neural ordinary differential equation. Because consistency follows from the transition function itself rather than from embedded physical quantities, the property is architecture-intrinsic and transfers across problem settings, as demonstrated in Yatkin et al. \cite{app15020933} on two independent datasets. Beyond satisfying the consistency and truncation requirements for sequential surrogates \cite{yatkin2026self}, ConsRNN demonstrated consistent damage initiation predictions under varying increment counts on both synthetic and FEM-generated data, including the elastoplastic benchmark of Bonatti and Mohr \cite{bonatti_2021_4642208}. Rather than serving as a replacement for empirical fracture criteria, ConsRNN introduces an additional learned information layer by compressing the localisation response of high-fidelity solid-element simulations into a fixed set of pre-trained network weights, learned offline rather than approximated at run time by a hand-calibrated criterion, thereby improving predictive accuracy while remaining compatible with arbitrary solver time discretisations. ConsRNN therefore provides a validated, consistency-preserving recurrent surrogate for history-dependent structural analysis.

Surrogate development, however, does not end with defining a mapping and training an architecture. An equally important challenge is deployment inside a finite element solver, where the trained network must execute incrementally while preserving its internal state across solver-driven time steps. Although recurrent surrogates for history-dependent material response have demonstrated promising offline predictive capability, comparatively little attention has been given to embedding customised recurrent architectures into production FE codes. Existing deployment approaches either rely on automated Keras-to-Fortran frameworks, which support only standard neural network layers \cite{PANTALE2022103647, DBLP:journals/corr/abs-2004-10652}, or hard-code neural network inference directly in Fortran \cite{10.1115/1.2389004, jang2021}. In the latter case, however, the embedded networks are memoryless, with path dependence supplied by the surrounding physics-based formulation rather than by the network itself. To the best of our knowledge, no previous study has embedded a customised, consistency-preserving recurrent architecture, whose non-standard transition function falls outside existing deployment frameworks, into a structural-scale finite element simulation.

The present paper closes this gap. We follow the hard-coding strategy and apply it to the Consistent Recurrent Neural Network (ConsRNN) \cite{app15020933}, previously validated only in Python/TensorFlow. The architecture is first optimised through Bayesian hyperparameter tuning, then reimplemented entirely in pure Fortran, with the learned weights transferred directly into an Abaqus VUMAT. As a comparison baseline we choose the SimpleRNN, which shares the same simple recurrent construction as ConsRNN without gating mechanisms, providing a more direct comparison than the gated LSTM and GRU architectures used by Bonatti and Mohr \cite{BONATTI2022104697}. We demonstrate that the consistency property of ConsRNN is preserved under solver-driven computation with varying incremental step sizes, whereas the SimpleRNN baseline exhibits increment-dependent behaviour and is therefore unsuitable for explicit finite element analysis. The embedded surrogate is further evaluated against the established two-parameter (2FS) fracture criterion, demonstrating that ConsRNN can serve as a state-carrying surrogate damage predictor for history-dependent structural simulations while introducing an additional learned information layer through knowledge transferred from high-fidelity solid-element simulations.

The remainder of the paper is organised as follows. Section~2 presents the optimisation of the ConsRNN and SimpleRNN architectures through Bayesian hyperparameter tuning. Section~3 describes the pure-Fortran implementation and the transfer of learned weights into the Abaqus VUMAT. Section~4 presents the numerical experiments, comparing the embedded ConsRNN with the SimpleRNN baseline and the 2FS fracture criterion. Section~5 concludes the paper and outlines future work.

\section{Neural Network Model Evaluation and Hyperparameter Tuning}
\label{sec3}

Before deployment within the FEM solver, the proposed neural network architectures are first developed and evaluated through a standard machine learning workflow. The workflow comprises data analysis, architecture design, Bayesian hyperparameter optimization, and comprehensive qualitative and quantitative performance evaluation to ensure a reproducible comparison between the candidate models. The following subsections describe each stage of this workflow and motivate the selection of the architecture used for the subsequent Fortran implementation.

\subsection{Path dependent damage prediction using bilinear loading dataset}
\label{sec:bilinear_dataset}
In an earlier study \cite{YATKIN_P1}, we used fine solid-element simulations to understand how in-plane deformation history affects damage evolution and subsequent localisation. A bilinear loading path was defined in terms of the principal major and minor strain and orientation angle, and a total of 19{,}098 distinct bilinear loading paths were simulated. Each path comprises two linear segments and an orientation angle, discretised into 400 increments. The simpleRNN and ConsRNN architectures (details in the next section) were trained on this dataset, split into 13{,}367 training, 2{,}866 validation, and 2{,}865 test paths. The validation set guided model selection during hyperparameter tuning, and final performance was reported on the held-out test set using the best-trained weights.

For further testing, 240 additional bilinear paths were generated and discretised at increment counts different from the original 400, which were used to assess both the consistency and the accuracy of the damage estimates, along individual loading paths and on the resulting forming limit curves. The bilinear-trained models were also evaluated in Python on more complex paths of three, four, and five linear segments. This test was motivated by the nature of structural simulations, where a single element may follow a complex, history-dependent loading path drawn from an effectively unbounded set of scenarios. Accordingly, two complementary evaluations were carried out. First, the trained models were embedded in the Abaqus/VUMAT framework to study their behaviour in structural simulations, and second, the bilinear-trained models were separately assessed on multilinear loading paths in Python.

\begin{equation}
\label{Convert}
\begin{aligned}
f(D_t) =& 1 - e^{-D_t}
\end{aligned}
\end{equation}

\begin{equation}
\label{Convert_inverse}
g(f(D_t)) = -\log(1 - f(D_t) + \epsilon)
\end{equation}
Here $\epsilon$ is a small numerical guard that keeps the logarithm bounded as $f(D_t)\to 1$.

\subsection{Architectures}
In the previous study~\cite{yatkin2026self}, we showed that conventional recurrent architectures such as GRU and LSTM fail to satisfy the consistency requirements of path-dependent deformation. Furthermore, their gating mechanisms complicate direct implementation within a fixed-form Fortran VUMAT, increase the computational cost, and make hard-coded deployment of a trained model more challenging. In contrast, the SimpleRNN and ConsRNN cells share the same single recurrent transition and similar architectural simplicity, while matching in depth, capacity, and the absence of gating mechanisms, making the SimpleRNN a natural baseline for evaluating the proposed consistency-preserving transition formulation. We therefore built two single-layer recurrent architectures based on the SimpleRNN and ConsRNN cells to enable a controlled comparison. The two models are not, however, an isolated test of the cell equation alone: the SimpleRNN ingests the accumulated strains together with the loading orientation $(\varepsilon_{1,t},\,\varepsilon_{2,t},\,\cot\phi_t)$, whereas the ConsRNN ingests the normalised strain increments. Consequently, the input representation, input dimensionality, and normalisation strategy also differ between the two architectures. The comparison should therefore be interpreted as an evaluation of two complete surrogate formulations rather than as an isolated comparison of the transition function itself. A single-layer SimpleRNN architecture also proved sufficient to learn the bilinear deformation paths over 400 increments at low computational cost. We therefore restricted both architectures to a single recurrent layer and tuned each through repeated training runs.

\begin{figure}[!h]
\centering
\includegraphics[scale = 0.9]{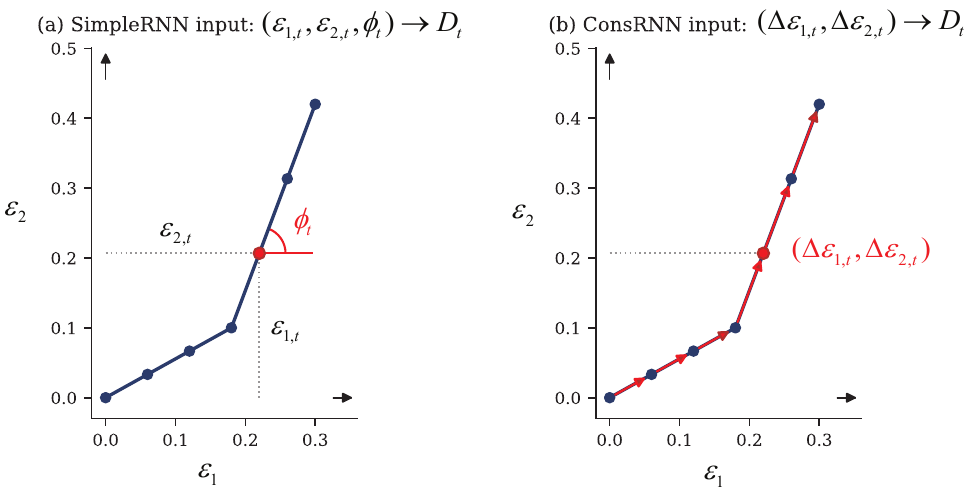}
\caption{Input representations of the same bilinear loading path for the two architectures in principal strain space. The red marker highlights the state of deformation at time $t$. Panel (a) shows the input to the SimpleRNN and panel (b) the input to the ConsRNN. Both panels depict the same underlying loading path, but the SimpleRNN encodes it through the accumulated strain state whereas the ConsRNN encodes it through the incremental strain differences.}
\label{fig:input_representation}
\end{figure}

The two cells differ in their inputs and in the transition functions used to update the state vectors. The SimpleRNN architecture takes the accumulated minor ($\varepsilon_{2,t}$) and major ($\varepsilon_{1,t}$) strains together with orientation ($\cot\phi_t = \Delta\varepsilon_{2,t}/\Delta\varepsilon_{1,t}$) at each time step $t$, as shown in Fig.~\ref{fig:input_representation} (a). The ConsRNN architecture instead takes the incremental strain differences ($\Delta\varepsilon_{1,t}$, $\Delta\varepsilon_{2,t}$) at each time step $t$, defined as the change in major and minor strain from the previous time step $t-1$ to the current time step $t$ as shown in Fig.~\ref{fig:input_representation} (b). In both cases the strains are principal plastic strains, and in the finite element implementation the network is advanced only on plastic increments: an elastic increment leaves the network state and the predicted damage unchanged. For both models, the target damage $D_t$ was mapped through the invertible function in Eq.~\eqref{Convert} during training, while predictions were mapped back to the original damage scale at inference using Eq.~\eqref{Convert_inverse}.

In Table~\ref{tab:ConsRNN_vs_SimpleRNN}, $X_t$ denotes the cell input at
time step $t$, that is, the strain increments $(\Delta\varepsilon_{1,t},\,
\Delta\varepsilon_{2,t})$ for ConsRNN and the accumulated strains with
orientation $(\varepsilon_{1,t},\, \varepsilon_{2,t},\, \phi_t)$ for
SimpleRNN. Further, $\hat{X}_t$ is its normalised form, $S_t$ the recurrent
state, and $A_t$ the transition activation obtained by applying the
$\tanh$ nonlinearity to the affine combination of input and previous
state. The matrices $W_x$ and $W_s$ are the trainable input and recurrent
weights, $W_{\mathrm{out}}$ the output weight, and $b$ the bias; these
parameters are learned during training rather than prescribed
analytically. Here $\|X_t\|_2$ denotes the Euclidean norm of the input
increment and $\sigma$ the sigmoid function.

\begin{table}[!htbp]
    \caption{Comparison of ConsRNN and SimpleRNN cells in terms of input transformation, normalisation, and state updates.}
    \centering
    \renewcommand{\arraystretch}{1.5}
    \begin{tabular}{lcc}
        \toprule
        \textbf{Step} & \textbf{ConsRNN} & \textbf{SimpleRNN} \\
        \midrule
        \textbf{Input} & $X_t = (\Delta \varepsilon_{1,t}, \Delta \varepsilon_{2,t})$ & $X_t = (\varepsilon_{1,t}, \varepsilon_{2,t}, \phi_t)$ \\
        \textbf{normalisation} & $\hat{X}_t = \frac{X_t}{\max(\|X_t\|_2, 10^{-15})}$ & - \\
        \textbf{Transition} & $A_t = \tanh(\hat{X}_t W_x + S_{t-1} W_s + b)$ & $S_t = \tanh(X_t W_x + S_{t-1} W_s + b)$ \\
        \textbf{Update} & $S_t = S_{t-1} + (A_t - S_{t-1}) \cdot \|X_t\|_2$ & $S_t = S_t$  \\
        \textbf{Output} & $\sigma(S_t \cdot W_{\text{out}} + b)$ & $\sigma(S_t \cdot W_{\text{out}} + b)$ \\
        \bottomrule
    \end{tabular}
    \label{tab:ConsRNN_vs_SimpleRNN}
\end{table}

\subsection{SimpleRNN Based Neural Network Architectures}

\begin{figure}[!h]
\centering
\includegraphics[scale = 0.6]{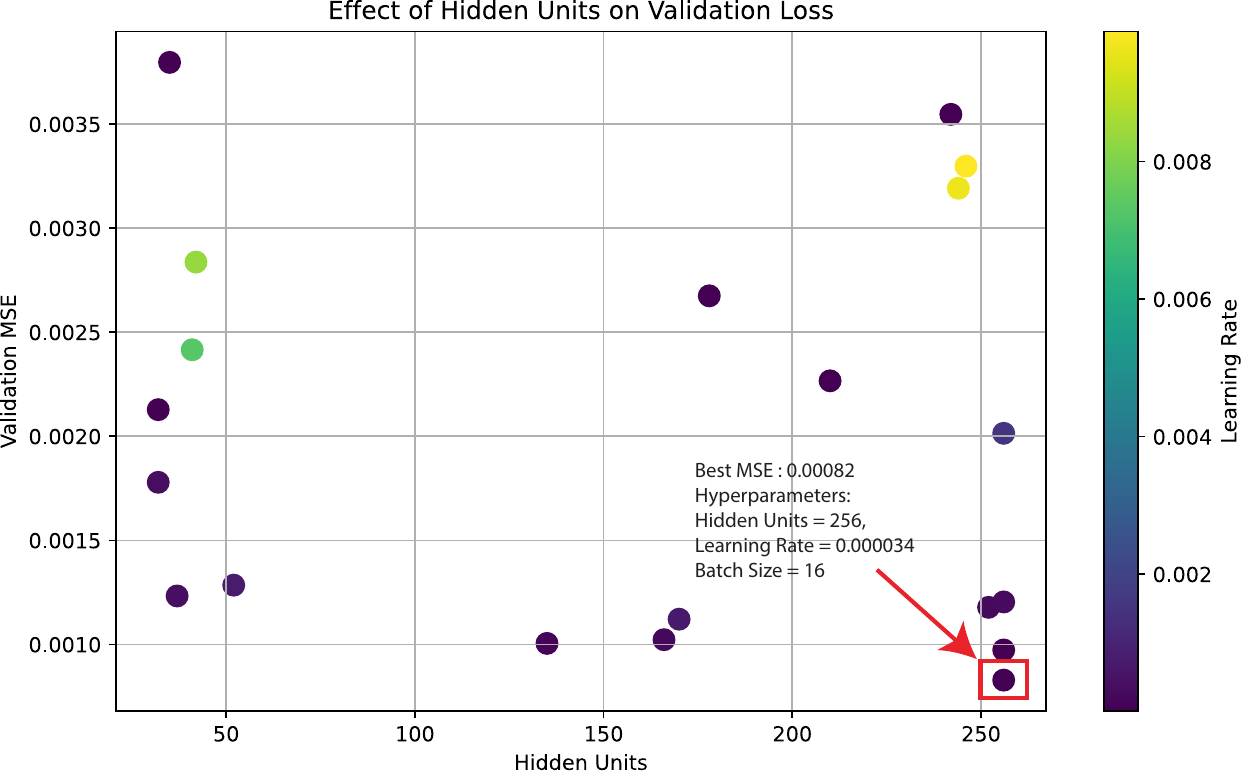}
\caption{Hyperparameter tuning of the SimpleRNN-based architecture over
20 Bayesian optimisation trials on the validation dataset. Each marker
denotes one trial, positioned by hidden units and validation MSE, coloured
by learning rate. The lowest MSE ($0.00082$) uses $256$ hidden units,
learning rate $3.4\times10^{-5}$, and batch size $16$ (highlighted).}\label{SimpleRNN_optimisation_Trials}
\end{figure}

The first architecture pairs the Keras/TensorFlow~\cite{tensorflow2015-whitepaper} SimpleRNN cell, whose update follows the standard formulation given in Table~\ref{tab:ConsRNN_vs_SimpleRNN}, with a single fully connected output layer using sigmoid activation. Its inputs are the accumulated minor and major strains $\varepsilon_{1,t}$ and $\varepsilon_{2,t}$ together with orientation $\cot\phi_{t}$, supplied at each time step through the mapping in Fig.~\ref{fig:input_representation} (a). Recall that, unlike ConsRNN, the cell operates on the accumulated strains rather than on their increments. Each candidate model was trained for 100 epochs, and the hyperparameters were tuned through Bayesian optimisation over 20 trials, jointly searching the learning rate, batch size, and number of hidden units; the trial-by-trial progression is reported in Fig.~\ref{SimpleRNN_optimisation_Trials}.

The best configuration uses 256 hidden units, a learning rate of $3.49 \times 10^{-5}$, and a batch size of 16, yielding a mean squared error of 0.0010 on the training set and 0.0008 on the validation set (Fig.~\ref{SimpleRNN_optimisation_Trials}). The close train--validation agreement indicates a well-conditioned fit; generalisation is assessed separately on the held-out test set (Table~\ref{tab:test_results}).

\subsection{ConsRNN Based Neural Network Architectures}
\label{subsec:consrnn_architectures}
The consistency property of the ConsRNN cell follows from its transition function, which differs from that of SimpleRNN cell, see Table~\ref{tab:ConsRNN_vs_SimpleRNN}. Rather than taking the accumulated strains directly, the cell operates on the strain increments between successive time steps: it forms a candidate state, subtracts the previous state, and scales the difference by the Euclidean norm of the input increment before applying the update. Because each state advance is proportional to the increment norm, the accumulated prediction over a fixed path converges as the increment size is refined rather than drifting with the number of increments a property we refer to as discretisation consistency: convergence of the prediction under refinement of the discretisation of a fixed strain path.

\begin{figure}[!h]
\centering
\includegraphics[scale = 0.6]{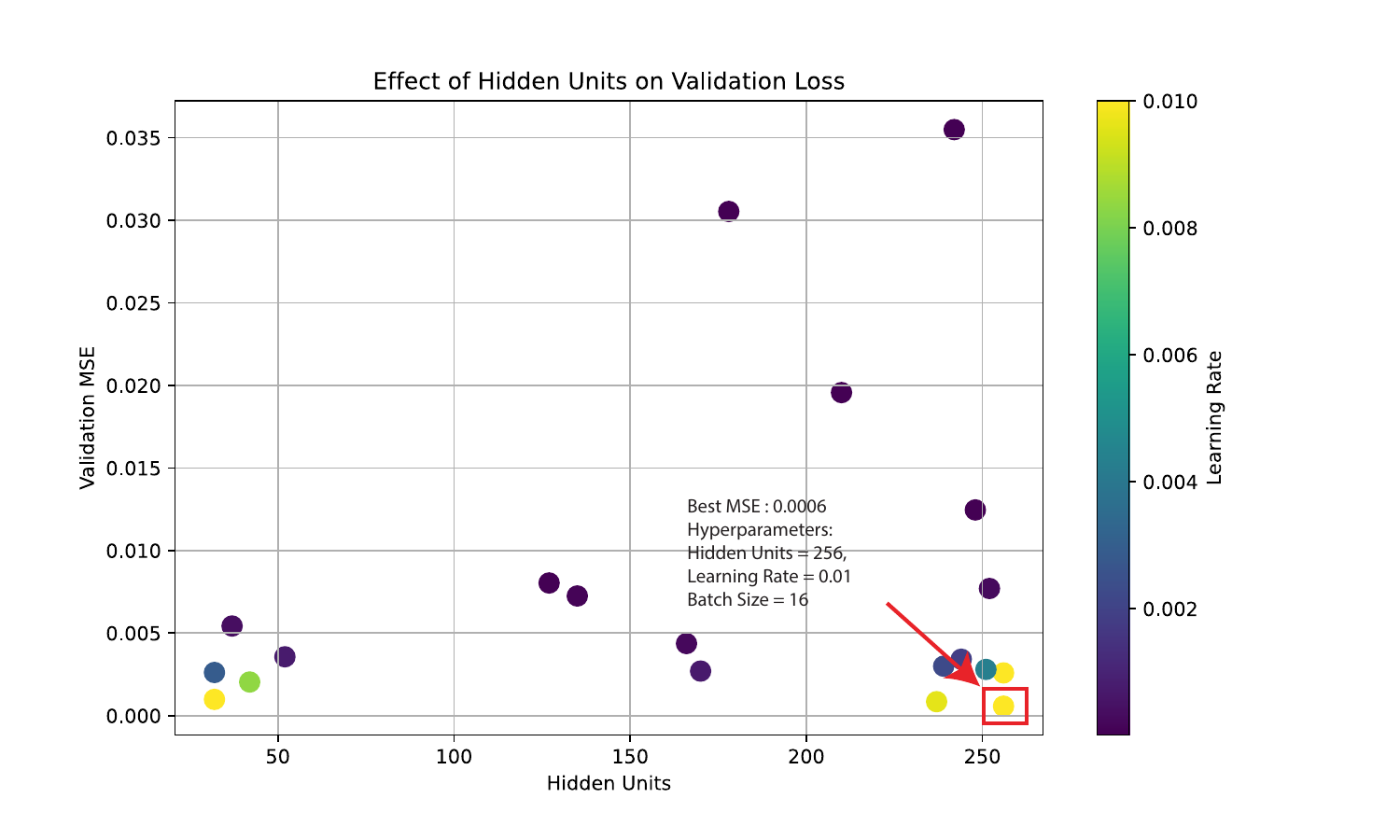}
\caption{Hyperparameter tuning of the ConsRNN-based architecture over
20 Bayesian optimisation trials on the validation dataset. Each marker
denotes one trial, positioned by hidden units and validation MSE, with
colour indicating the learning rate. The lowest MSE ($0.0006$) is obtained
with $256$ hidden units, learning rate $0.01$, and batch size $16$
(highlighted); batch size $256$ was selected for the study, as explained
in the text.}\label{ConsRNN_optimisations}
\end{figure}

The second architecture replaces the SimpleRNN cell with the ConsRNN cell defined in Table~\ref{tab:ConsRNN_vs_SimpleRNN}, followed by a single fully connected output layer using sigmoid activation. Its inputs are the strain increments $\Delta\varepsilon_{1,t}$ and $\Delta\varepsilon_{2,t}$ between successive time steps, supplied through the mapping in Fig.  \ref{fig:input_representation}(b). Following the same protocol as for the SimpleRNN, each candidate model was trained for 100 epochs and the hyperparameters were tuned through Bayesian optimisation over 20 trials, jointly searching the number of hidden units in $[32, 256]$, the learning rate on a log-uniform scale over $[10^{-5}, 10^{-2}]$, and the batch size in $[16, 256]$; the trial-by-trial progression is reported in Fig.~\ref{ConsRNN_optimisations}.

The best configuration uses 256 hidden units, a learning rate of $1.0\times 10^{-2}$, and a batch size of 16, yielding a mean squared error of 0.0006 on both the training and validation sets (Fig.~\ref{ConsRNN_optimisations}). As with the SimpleRNN, the close train--validation agreement indicates a well-conditioned fit; generalisation is assessed separately on the held-out test set (Table~\ref{tab:test_results}).

Although the lowest validation MSE was obtained with a batch size of 16, this configuration showed reduced accuracy at increment counts away from the 400-increment training resolution, while still preserving discretisation consistency. A configuration with batch size 256 achieved comparable consistency together with higher accuracy across all tested increment counts, and was therefore selected for the results reported in the remainder of this study.


\begin{figure}[t]
\centering
\includegraphics[scale = 0.7]{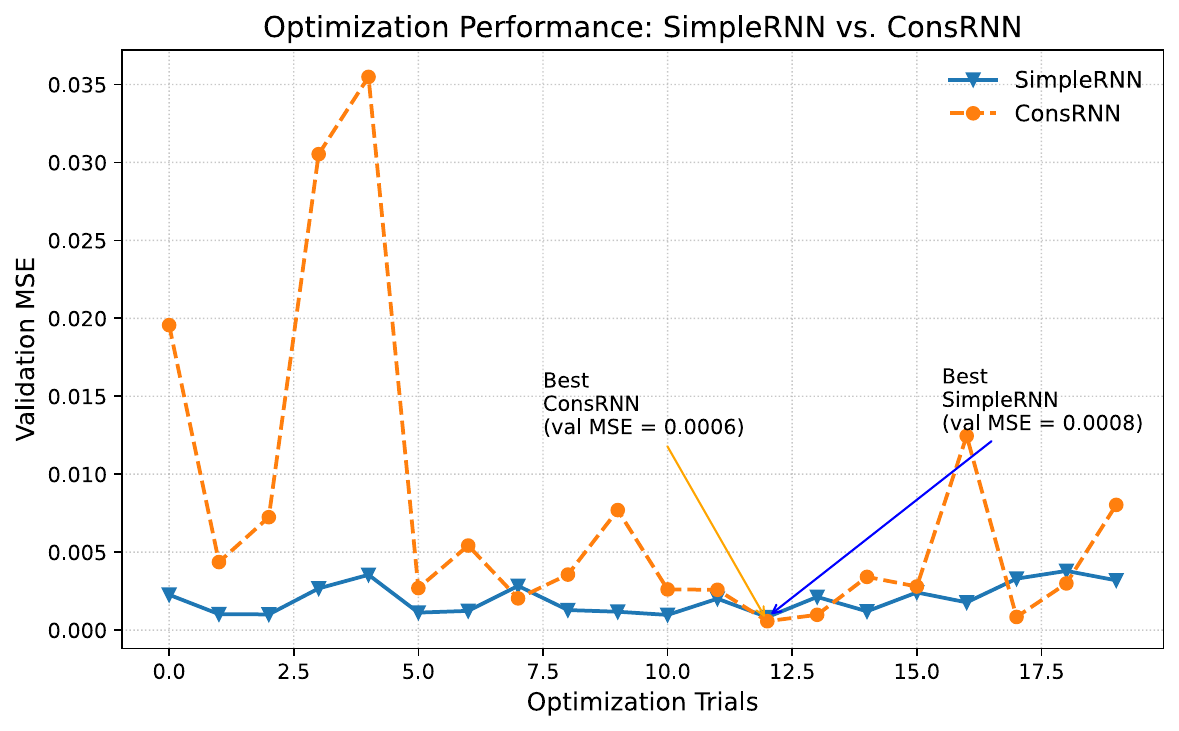}
\caption{Validation mean squared error (MSE) across the 20 Bayesian optimisation trials for the SimpleRNN and ConsRNN architectures. The SimpleRNN attains consistently low MSE across trials, whereas the ConsRNN exhibits larger trial-to-trial variability; nevertheless, the best-performing configuration overall is a ConsRNN model, reaching the lowest validation MSE of the search (0.0006 versus 0.0008 for the best SimpleRNN).}\label{optimisation_Trials}
\end{figure}

\subsection{Held-Out Test-Set Performance}
The held-out test set results, summarized in Table~\ref{tab:test_results},
are consistent with the hyperparameter optimisation results on the
bilinear loading dataset: the lowest-validation-MSE ConsRNN architecture
outperforms the corresponding SimpleRNN architecture on every metric and
on both output
scales. On the converted scale, ConsRNN reduces the RMSE from $0.0281$
to $0.0236$ and the MAE from $0.0188$ to $0.0150$, and raises the
coefficient of determination from $R^2 = 0.9832$ to $R^2 = 0.9882$.
The same ordering holds on the original damage scale, where ConsRNN
lowers the RMSE from $0.0616$ to $0.0440$ and improves $R^2$ from
$0.9784$ to $0.9890$.

The advantage extends to the tail of the error distribution:
the $95$th-percentile absolute error, which characterizes worst-case
behaviour across the test paths, is also lower for ConsRNN on both
scales ($0.0502$ versus $0.0586$ on the converted scale, and $0.0869$
versus $0.1161$ on the damage scale). ConsRNN therefore fits the typical
paths more accurately while remaining more reliable on the difficult
cases. When the lowest-validation-MSE model of each architecture is
selected and evaluated once on the untouched test set, ConsRNN
is the more accurate surrogate of the two.

These results characterise the two architectures at their respective
validation optima. As noted in Section~\ref{subsec:consrnn_architectures},
the ConsRNN configuration carried forward to the Fortran implementation
and all subsequent experiments uses batch size 256, selected for its
accuracy across increment counts rather than for its validation MSE at
the training resolution.

\begin{table}[ht]
\centering
\caption{Test set results for SimpleRNN and ConsRNN on both output scales. Both are the lowest-validation-MSE configurations (batch size 16).}
\label{tab:test_results}
\begin{tabular}{llcccccc}
\toprule
\textbf{Output Type} & \textbf{Model} & \textbf{RMSE} & \textbf{MAE} & \textbf{nRMSE} & \textbf{MAPE (\%)} & \textbf{P95 abs err} & \textbf{$R^2$} \\
\midrule
Converted Output & SimpleRNN & 0.0281 & 0.0188 & 0.0289 & 6.23 & 0.0586 & 0.9832 \\
               & ConsRNN    & 0.0236 & 0.0150 & 0.0243 & 5.40 & 0.0502 & 0.9882 \\
\addlinespace
Original Damage Output & SimpleRNN & 0.0616 & 0.0378 & 0.0174 & 7.92 & 0.1161 & 0.9784 \\
               & ConsRNN    & 0.0440 & 0.0281 & 0.0124 & 6.71 & 0.0869 & 0.9890 \\
\bottomrule
\end{tabular}
\end{table}

\subsection{Localisation Predictions for Bilinear Paths}
\begin{figure}[H]
\centering
\includegraphics[scale = 0.8]{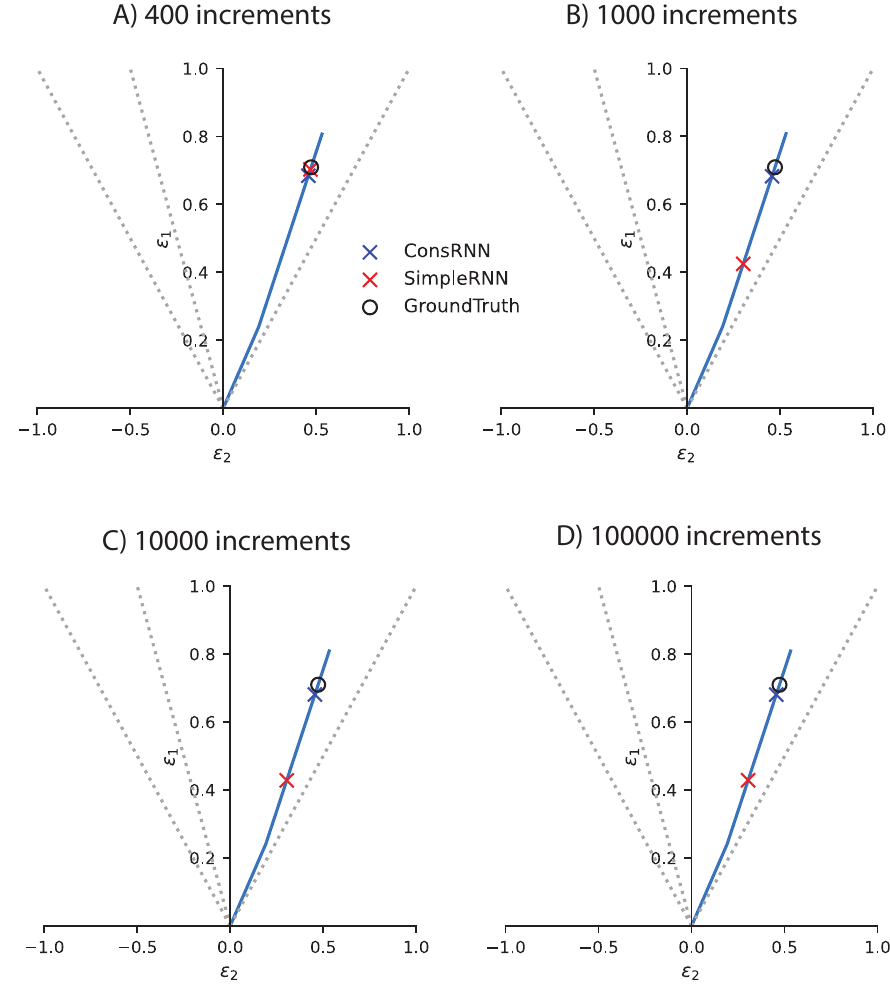}
\caption{Localisation predictions obtained from SimpleRNN and ConsRNN are
compared under different numbers of increments. While increasing the number
of increments causes SimpleRNN predictions to drift, ConsRNN predictions
remain consistent.}\label{Predictions_Incremental_Change}
\end{figure}

To examine how the predicted localisation points depend on temporal
resolution, we evaluated both trained architectures in Python/TensorFlow
along a single bilinear deformation path, discretised at $400$, $1{,}000$,
$10{,}000$, and $100{,}000$ increments
(Fig.~\ref{Predictions_Incremental_Change}). The path consists of a first
segment of length $R_1 = 0.30$ with direction $\cot\phi_1 = 0.80$ and a
second segment of length $R_2 = 0.67$ with $\cot\phi_2 = 0.60$. The
ground-truth localisation point, marked by the empty circle in each panel, is
obtained directly from the finite element simulation of this path, in which
localisation occurs at increment $84{,}426$ of $100{,}001$. The path is held
fixed across all four panels, so that the only quantity changing between them
is the number of increments used to traverse the identical strain history.

As the increment count increases, the SimpleRNN predictions drift
progressively away from the ground-truth localisation point: at the
training resolution of $400$ increments the prediction is accurate, but at higher increment counts it departs visibly from the reference. The ConsRNN predictions, in contrast, remain close to and converge as the resolution is refined. This drift-free behaviour reflects the discretisation consistency built into the ConsRNN transition function: because each state advance is scaled by the norm of the strain increment, the accumulated prediction converges under refinement toward a value set by the path itself rather than by how finely it is sampled. It is this property that makes the surrogate usable across the varying, solver-dictated step sizes encountered in FE simulations, where the increment count cannot be prescribed by the user. The two architectures also differ in their input representation, not only in the transition function, so this contrast reflects the two complete designs rather than the transition function in isolation.

\subsection{Forming Limit Curve (FLC) Predictions}
\label{flc_predictions}
\begin{figure}[H]
\centering
\includegraphics[scale = 0.8]{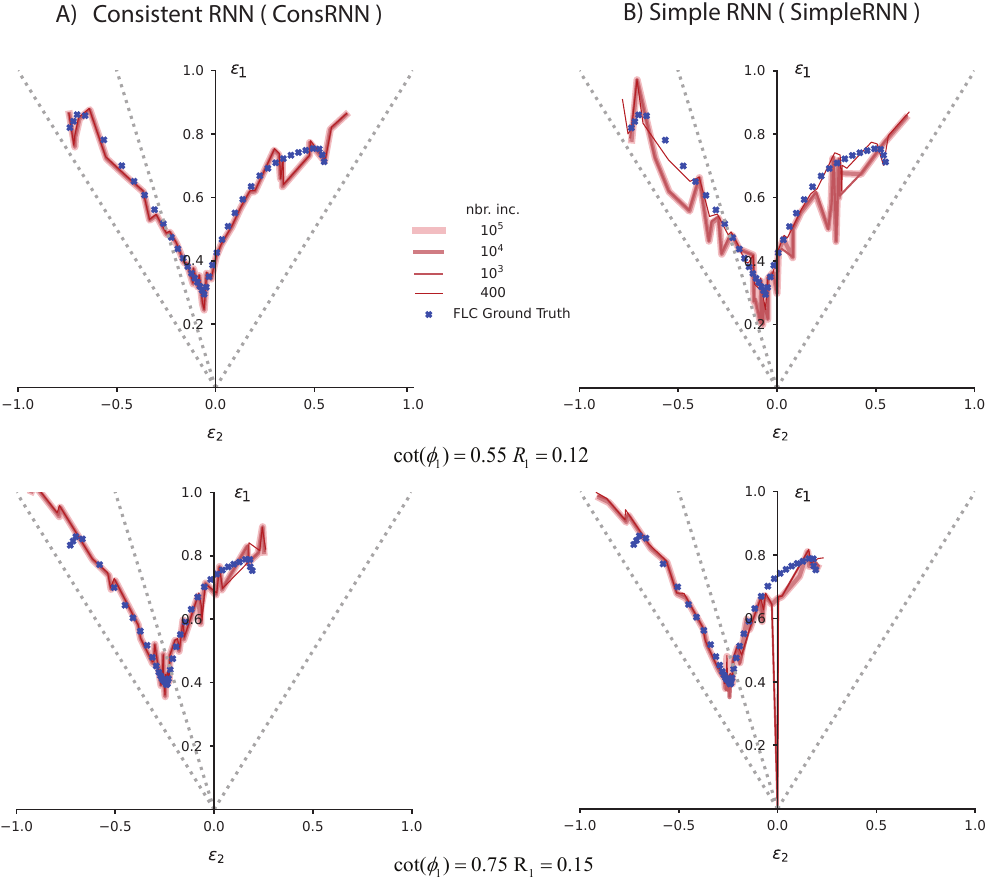}
\caption{FLC predictions from SimpleRNN and ConsRNN trained neural network architectures.}\label{SimpleRNN_ConsRNN_FLC_comparison}
\end{figure}

The two selected architectures, the deployed ConsRNN (batch size 256) and the SimpleRNN, were used to predict localisation points that define two Forming Limit Curves (FLCs), shown in Fig.~\ref{SimpleRNN_ConsRNN_FLC_comparison}. Each FLC is traced by fixing the first linear segment and sweeping the direction of the second segment: the first FLC fixes the first segment at length $R_1 = 0.12$ with direction $\cot\phi_1 = 0.55$, and the second at $R_1 = 0.15$ with $\cot\phi_1 = 0.75$. In both cases the second segment is held at fixed length while its direction $\cot\phi_2$ is varied across $40$ values, each value yielding one point on the curve. Here $\cot\phi = \Delta\varepsilon_2/\Delta\varepsilon_1$ denotes the ratio of the minor- to major-strain increment, following the convention used throughout this work. Each path was simulated at $100{,}000$ increments and then downsampled to $10{,}000$, $1{,}000$, and $400$ increments for prediction, so that the FLCs probe both accuracy and discretisation consistency. The ground-truth localisation points were obtained from Marciniak--Kuczy\'nski finite element simulations in Abaqus of these bilinear paths, following the dataset described in Section~\ref{sec:bilinear_dataset}, with the localisation strain read directly from each simulated path. These reference localisations are plotted alongside the SimpleRNN and ConsRNN predictions.

As Fig. \ref{SimpleRNN_ConsRNN_FLC_comparison} shows, the ConsRNN-based architecture outperforms the SimpleRNN in both accuracy and consistency. It is nonetheless worth noting that the SimpleRNN predictions are far more consistent than those of the GRU model examined in the previous study \cite{app15020933}. This difference comes mainly from the theoretical structure of the GRU transition function, which includes a gating mechanism \cite{DBLP:journals/corr/DeyS17}. While this mechanism improves learning ability and accuracy, it also reduces the consistency of the estimations, as observed in the results in the previous study \cite{app15020933}.

\subsection{Localisation Predictions for Multilinear Paths}
\label{Predictions_in_python}

\begin{figure}[t]
\centering
\includegraphics[scale = 0.9]{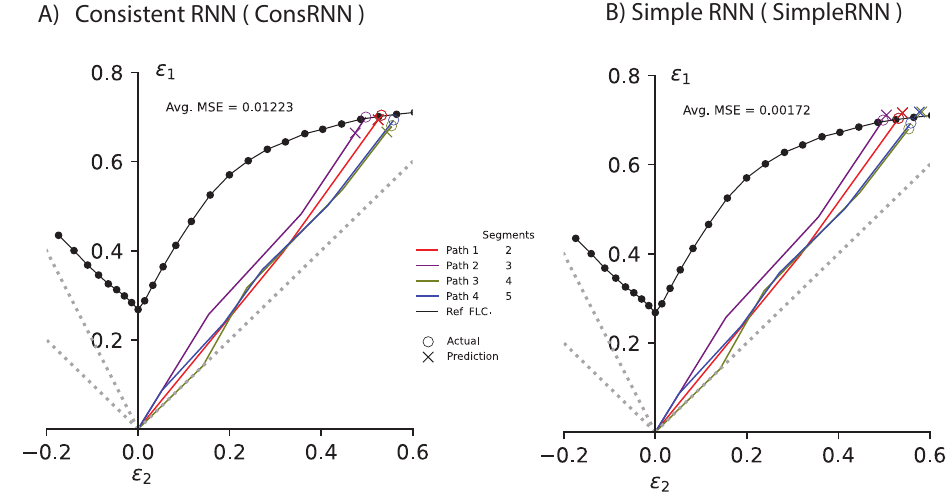}
\caption{Localisation predictions for multilinear paths (two to five segments)
from the deployed ConsRNN (A, batch size 256) and the SimpleRNN (B)
architectures over 400 increments. The black curve with filled markers is the
reference forming limit curve from Fig.~\ref{SimpleRNN_ConsRNN_FLC_comparison}. Circles:
ground truth; crosses: predictions.}\label{SimpleRNN_ConsRNN_multilinear_paths_comparison}
\end{figure}

Although both architectures were trained only on bilinear loading paths, the mappings in Fig. \ref{fig:input_representation} (a) and (b) are history dependent and can therefore be applied to more complex multilinear paths. To test this, we generated multilinear paths of two, three, four, and five segments at 400 increments and predicted their localisation points with the bilinear-trained models. Each segment is specified by a length $R$ and a direction expressed through $\cot\phi_{t} = \Delta\varepsilon_{2,t}/\Delta\varepsilon_{1,t}$, the ratio of the minor- to major-strain increment. The three-segment path is defined by the sequence $(R,\,\cot\phi_{t}) = (0.30,\,0.60),\,(0.30,\,0.90),\,(0.30,\,0.65)$; the four-segment path by $(0.20,\,1.00),\,(0.20,\,0.55),\,(0.30,\,0.95),\,(0.30,\,0.75)$; and the five-segment path by $(0.10,\,0.60),\,(0.20,\,0.90),\,(0.15,\,0.70),\,(0.20,\,1.00),\,(0.30,\,0.75)$. Both architectures predict the localisation points as shown in Fig.~\ref{SimpleRNN_ConsRNN_multilinear_paths_comparison}.

On these multilinear paths the SimpleRNN is the more accurate of the two, with an average MSE of $0.00172$ versus $0.01223$ for the ConsRNN (Fig.~\ref{SimpleRNN_ConsRNN_multilinear_paths_comparison}). The ConsRNN evaluated here is the deployed batch-256 configuration carried forward to the Fortran implementation, rather than the lowest-validation-MSE batch-16 configuration reported in Table~\ref{tab:test_results}; as noted in Section~\ref{subsec:consrnn_architectures}, it was selected for accuracy across increment counts rather than at the 400-increment training resolution. At this fixed resolution the standard state update of the SimpleRNN is free to fit the training discretisation more closely. The results of this controlled out-of-distribution test reflect a genuine trade-off: at a single fixed increment count matching the training resolution, the convex, norm-scaled state update of the ConsRNN sacrifices accuracy to guarantee discretisation consistency, which the SimpleRNN does not provide. For deployment inside an Abaqus/VUMAT surrogate this trade-off favours the ConsRNN, because the solver assigns step sizes that can vary by orders of magnitude across elements and across increments. A SimpleRNN evaluated at resolutions away from its training increment count drifts from the correct localisation (Fig.~\ref{Predictions_Incremental_Change}), and its fixed-resolution accuracy edge does not survive this variability, whereas the ConsRNN preserves both its consistency and its accuracy as the increment count increases. The ConsRNN is therefore the appropriate choice for FEM embedding, while the SimpleRNN remains competitive only where the increment count is fixed at the training resolution and known in advance.

\section{Fortran Implementation}
\label{impmementation_methodology}
Neural networks are typically built and trained in TensorFlow or PyTorch, whereas the finite-element simulations in this work are run in Abaqus, where the constitutive response is supplied through Fortran user subroutines. Bridging the two ecosystems is not straightforward. Frameworks such as Neural-Fortran \cite{curcic2019parallelfortranframeworkneural} and PyTorch-ABAQUS UMAT \cite{SUH2023104682} have been proposed to couple TensorFlow or PyTorch with Fortran, but they are built around standard Keras and PyTorch layers, such as convolutional and recurrent layers. The Consistent Recurrent Neural Network (ConsRNN) developed in our previous work \cite{app15020933} uses a non-standard state-update mechanism that these bridging frameworks cannot represent directly.

\begin{figure}[!h]
\centering
\includegraphics[scale = 0.8]{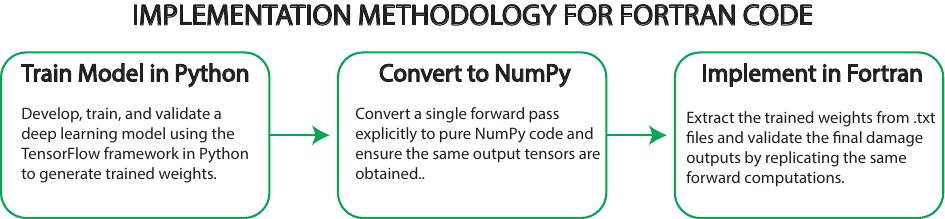}
\caption{Three-stage workflow for the Fortran implementation of the ConsRNN architecture: training in Python/TensorFlow, replication in NumPy, and rewriting in pure Fortran, validated against each other at every stage. The resulting code is embedded into the Abaqus VUMAT subroutine (Section~\ref{impmementation_methodology}).}\label{fig1}
\end{figure}

We therefore reimplemented ConsRNN from scratch, reconstructing each computational block normalisation, state transition, state update, and damage output in native code. A similar block-by-block strategy has been adopted in related work \cite{PANTALE2022103647, 10.1115/1.2389004}. This route gives full control over both the model and its numerical evaluation inside the solver. The workflow proceeds in three stages. The models are first trained in TensorFlow, and the trained weights are exported to text files. The forward pass is then replicated in NumPy from these weight files and matched against the TensorFlow output. Finally, each block is rewritten in pure Fortran, with the tensor outputs validated against the NumPy reference at every step shown in Fig.~\ref{fig1}. The result is a self-contained Fortran implementation of ConsRNN that integrates directly into Abaqus FEM simulations.

\subsection{Python to Abaqus/Vumat Fortran} 

The incremental update mechanism of RNNs, in which a state vector is carried as memory across time steps, makes them well suited to material modelling, as emphasised by Mohr and co-workers \cite{doi:10.1126/sciadv.abf3658, HEIDENREICH2024116991, BONATTI2022104697}. Although architectures such as Transformers and CNNs can also represent deformation history, the recurrent update offers a decisive advantage for this purpose by advancing the state one increment at a time without storing or growing a sequence. This matches the structure of an explicit finite-element solver exactly. At each increment the RNN applies a transition function and then computes the output from the updated state, which is the same order of operations Abaqus performs at each material point. We therefore implemented the simulation environment from scratch in Python and translated the same logic into Fortran for embedding in the Abaqus VUMAT.

The trained SimpleRNN and ConsRNN one-layer architectures were first
converted into pure scratch calculations, in which the damage predictions
were matched between the TensorFlow model outputs and the NumPy-based
scratch implementations. The same scratch implementations were
subsequently rewritten in pure Fortran using the same inputs and tensor
outputs.

The forward passes of the SimpleRNN- and ConsRNN-based architectures are
summarized as pseudocode in
Algorithms~\ref{alg:simplernn}~and~\ref{alg:consrnn}
(Appendix~\ref{app:pseudocode}), and the complete Python and Fortran
implementations are openly available in the accompanying repository (see the Data Availability section). For both architectures,
the Python and Fortran implementations produced identical damage outputs
on the same input data, matching to five decimal places.

This cross-language validation established that the Fortran routines faithfully reproduce the trained networks. Only ConsRNN, however, was subsequently embedded in the Abaqus/VUMAT environment for the numerical experiments reported in this section. The earlier TensorFlow experiments in Section~\ref{flc_predictions} showed that SimpleRNN does not satisfy the discretisation-consistency property: its predicted damage trajectory varies with the increment count even when
the underlying strain path is identical. In an explicit finite-element simulation the
increment count is dictated by the stable time step and can vary by
orders of magnitude between runs, so a network with this property would yield damage predictions governed by the solver's time-stepping rather than the physical deformation history. We therefore restricted the FEM-coupled 
experiments to ConsRNN,  which is discretisation-consistent by construction. 

The ConsRNN implementation is structured so that, once the simulation
reaches a prescribed sampling frequency, the ConsRNN memory is updated
by preserving the hidden-state vector within the \texttt{StateEnv}
variables of the subroutine. The incremental differences between
successive sampling instants are also stored, after which the update
mechanism is applied and the damage is computed at each prescribed
sampling point.

At each sampling instant the
ConsRNN state vector is updated using the transition function presented
in Table~\ref{tab:ConsRNN_vs_SimpleRNN}, and the corresponding damage
value is computed. Crucially, no growing incremental
input sequence of minor and major strain values is assembled and
passed to the network, as is typical in TensorFlow-based
implementations. The state vector is updated in place, the damage is computed immediately, and both are stored in the \texttt{StateEnv} variables.

In this way, the ConsRNN state vector and the damage evolution are
consistently preserved and updated through the \texttt{StateEnv}
structure across sampling instants. This allows ConsRNN to act as an
FEM-compatible incremental solver while remaining memory-efficient and
fully aligned with the way finite-element simulations operate in
practice. This correspondence is the central objective of the present
implementation: it enables ConsRNN to function as a physically
consistent surrogate model that naturally follows the incremental
strain evolution produced by the FEM simulation.

\section{Numerical Experiments}
\label{sec4}

Reproducing a forward pass in Fortran is not the same as operating inside a
solver, where the network is driven by solver-chosen increments at a resolution
that varies between elements and between runs, along strain paths never seen
during training. Deployment also exposes behaviour that offline testing cannot
reveal, since the subroutine must carry its state through the solver's own
memory management across thousands of material points without disrupting the
analysis. The embedded surrogate is therefore evaluated in three stages:
prescribed bilinear paths in single-element simulations, where the increment
count is varied while the path is held fixed; element-level damage predictions
in a full structural simulation, where each element follows its own
non-proportional history; and finally the global force-displacement response
with the surrogate governing element deletion, compared against the established
2FS criterion.

\subsection{Single Element Simulations}
\begin{figure}[!h]
\centering
\includegraphics[scale = 0.9]{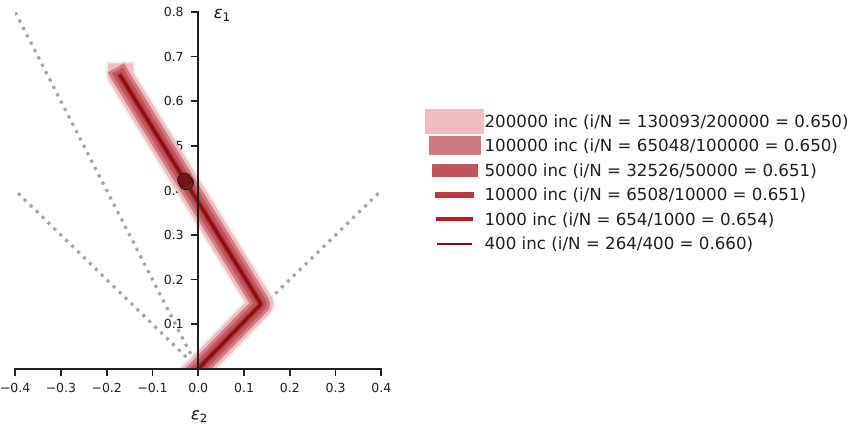}
\caption{Localisation predictions of ConsRNN embedded in Abaqus/Explicit for the first bilinear strain path ($R_{1,2} = 0.2,\, 0.6$; $\cot(\phi_{1,2}) = 0.95,\, -0.6$), evaluated at increment counts from 400 to 200{,}000. Markers denote the predicted localisation point.}
\label{fig:path1}

\end{figure}

\begin{figure}[!h]
\centering
\includegraphics[scale = 0.9]{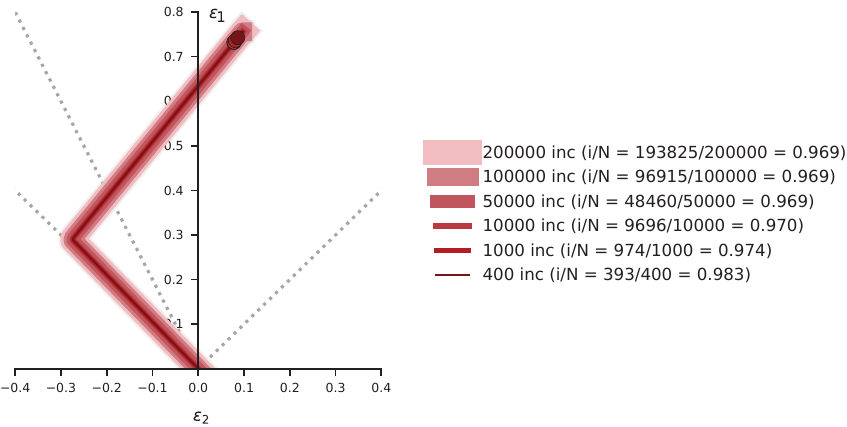}
\caption{Localisation predictions of ConsRNN embedded in Abaqus/Explicit for the second bilinear strain path ($R_{1,2} = 0.4,\, 0.6$; $\cot(\phi_{1,2}) = -0.95,\, 0.80$), evaluated at increment counts from 400 to 200{,}000. Markers denote the predicted localisation point.}
\label{fig:path3}
\end{figure}

Two different bilinear loading paths, previously introduced in \cite{korgesaar2025machine}, have been simulated in Abaqus using the VUMAT subroutine. The Fortran implementations of ConsRNN were tested by varying the loading frequencies. For each loading path, the strain increments extracted from the simulations were set to 400, 1000, 10000, 50000, 100000, and 200000. The corresponding results for each path are shown in Fig.~\ref{fig:path1} and Fig.~\ref{fig:path3}. In addition, the localisation index ratio as a function of the number of increments is also included in the figures.

Each bilinear path consists of two linear segments in principal strain space, where each segment $i$ is defined by an arc length $R_i$ and a direction expressed through $\cot\phi_i = \Delta\varepsilon_{II}/\Delta\varepsilon_{I}$, the ratio of the minor- to major-strain increment, following the convention used for the multilinear paths in Section~2.8. Step time is allocated to the segments in proportion to their lengths, such that the path is traversed at constant speed in strain space. The first path, shown in Fig.~\ref{fig:path1}, is defined by $R_{1,2} = 0.2,\, 0.6$ and $\cot(\phi_{1,2}) = 0.95,\, -0.6$. Its first segment ($\cot\phi_1 = 0.95$) corresponds to nearly equibiaxial stretching, reaching $(\varepsilon_{II}, \varepsilon_{I}) = (0.138,\, 0.145)$, after which the loading direction changes abruptly by $74.5^{\circ}$ into the second segment ($\cot\phi_2 = -0.6$), which drives the material toward the drawing side of the strain plane. The predicted localisation instant, expressed as the normalised increment ratio $i/N$ at which the damage indicator reaches unity, converges monotonically from $0.660$ at $N = 400$ to $0.650$ for $N \geq 10^{5}$, a spread of $1.5\%$ across a 500-fold range of temporal resolution.

The second path, shown in Fig.~\ref{fig:path3}, is defined by $R_{1,2} = 0.4,\, 0.6$ and $\cot(\phi_{1,2}) = -0.95,\, 0.80$, and represents the reverse scenario. Its first segment ($\cot\phi_1 = -0.95$) lies close to pure shear, driving the material deep into the drawing side of the strain plane and reaching $(\varepsilon_{II}, \varepsilon_{I}) = (-0.276,\, 0.290)$, after which the loading direction changes by $82.2^{\circ}$ into the second segment ($\cot\phi_2 = 0.80$), which pulls the material back toward nearly equibiaxial stretching. Along this path the damage accumulates more slowly, and localisation is reached only near the end of the prescribed loading: the predicted ratio converges monotonically from $0.983$ at $N = 400$ to $0.969$ for $N \geq 5 \times 10^{4}$, a spread of $1.4\%$, mirroring the consistency observed for the first path.

As can be seen from the results, although the predictions show a minor deviation for the coarsest case with 400 increments, ConsRNN consistently preserves the localisation point as the number of increments increases, regardless of whether localisation occurs in the middle of the deformation path or near its end. This behavior indicates that ConsRNN is well suited for incremental material evolution: its memory mechanism carries the accumulated damage information between increments, so the prediction depends on the loading path itself rather than on its temporal discretisation, effectively acting as an incremental solver.

\subsection{Structural simulation and Material Model}
In contrast to single shell element simulations where the prescribed in-plane loading was applied to the element, the elements in actual structures are exposed to infinitely complex loading histories. Therefore, the goal of this section is to reveal how the implemented ML models behave under these conditions. 


The structural model used in this analysis is a rectangular steel plate
clamped along all of its boundaries and deformed by a prescribed
out-of-plane displacement applied through a rigid indenter at its centre.
The plate is discretised with four-node reduced-integration shell elements
of 3 mm thickness, with three Gauss integration points through the thickness
so that both membrane and bending response are captured. The analysis is carried
out in Abaqus/Explicit, and the constitutive behaviour is supplied through
a user-defined material subroutine (VUMAT). This setup exposes the elements
to non-proportional, spatially varying loading histories that differ from
the prescribed single-element paths used in the previous section, since the
strain history experienced by each element depends on its position relative
to the indenter and on the evolving deformation of the surrounding material.

The same material model is used as in the previous section. The steel is
described by a rate-independent, isotropic elastic-plastic constitutive law
formulated under the plane-stress assumption appropriate for shell elements.
The elastic response is linear and isotropic, defined by the Young's modulus
$E$ and the Poisson's ratio $\nu$. Plastic flow follows the von Mises (J2)
yield condition with associated flow rule and isotropic hardening. The flow
stress obeys a modified Swift hardening rule, in which the stress is held at
the initial yield stress $\sigma_0$ up to a L\"uders plateau strain
$\varepsilon_L$, beyond which it follows the power-law relation
\begin{equation}
\sigma_f = K \left( \varepsilon_p + \varepsilon_0 \right)^{n},
\end{equation}
where $\varepsilon_p$ is the equivalent plastic strain, $K$ and $n$ are the
hardening parameters, and the offset strain
$\varepsilon_0 = (\sigma_0/K)^{1/n} - \varepsilon_L$ enforces continuity of the
stress-strain curve at $\varepsilon_p = \varepsilon_L$. The constitutive
equations are integrated using a plane-stress-consistent return-mapping
algorithm, in which the through-thickness strain component is recovered
iteratively so that the zero normal-stress condition is enforced at every
increment. The steel is modelled with $E = 200$ GPa, $\nu = 0.3$, density
7850 kg/m$^3$, $\sigma_0 = 272$ MPa, $K = 620$ MPa, and $n = 0.2$.

Ductile fracture is modelled with the two-factor-scaling (2FS) criterion
\cite{korgesaar2019effect}, in which the equivalent fracture strain is interpolated
between an upper bound given by the Modified Mohr--Coulomb (MMC) fracture locus
and a lower bound given by the Swift diffuse-necking strain, according to the
ratio of the characteristic element length to the plate thickness. This
two-factor scaling regularises the mesh dependence of strain localisation in
coarsely meshed shell structures, since the localisation and fracture strains
exhibit different element-size sensitivities. Damage is accumulated through a
linear rule as the integral of the equivalent plastic strain increment
normalised by the current fracture strain, and fracture occurs when the
accumulated damage reaches unity, at which point the element is removed from the
mesh through the standard element-deletion procedure. Following the original formulation, a triaxiality cut-off of $\eta = 1/3$ is applied, below which fracture is not predicted, since the Swift necking lower bound is defined only for $1/3 \leq \eta \leq 2/3$.

\subsection{ConsRNN based Damage Predictions in Structural Simulation}
In a structural simulation, the number of strain increments accumulated by an
element throughout its deformation history depends on the element location, the
local deformation severity, and the adaptive timestep strategy of the solver.
As a result, the number of increments can vary substantially between different
elements and also between integration points within the same element. For a
surrogate damage model to be practically useful within a simulation
environment, its predictions should remain insensitive to such
temporal-resolution variability.

\begin{figure}[!h]

\centering
\includegraphics[scale = 0.9]{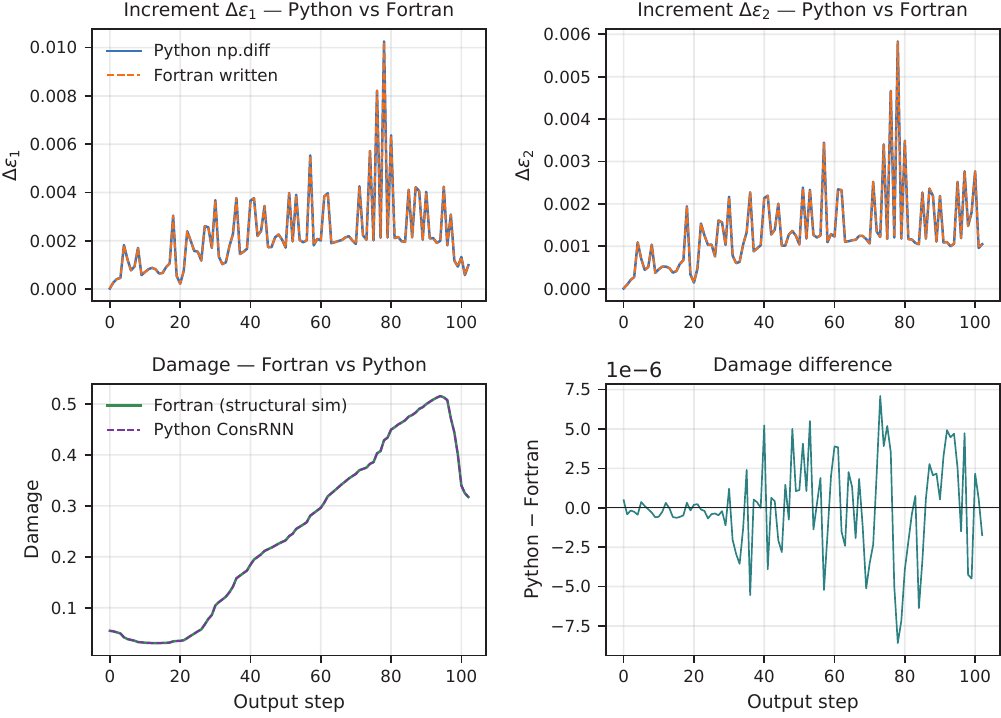}
\caption{Validation of the embedded ConsRNN at Element~2852, Integration
Point~2 of the structural simulation: Fortran-computed strain increments and
damage evolution against the Python reference.}
    \label{fig:consrnn_vumat_validation}
\end{figure}

\begin{figure}[!h]
\centering
\includegraphics[scale = 0.9]{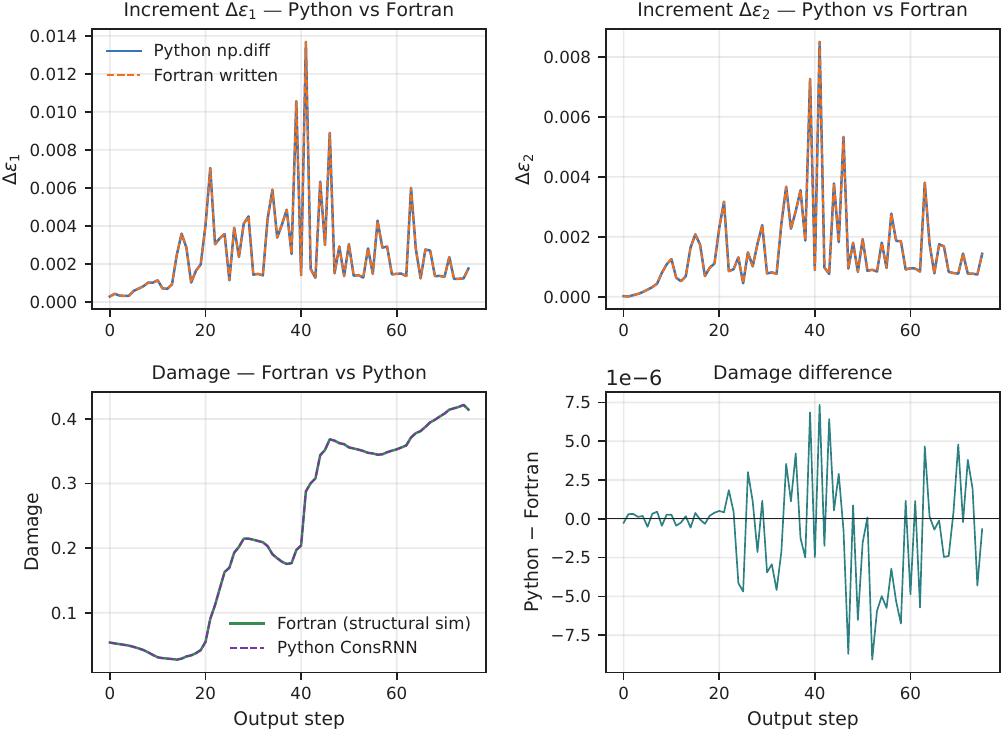}
\caption{Validation of the embedded ConsRNN at Element~3089, Integration
Point~3 of the structural simulation: Fortran-computed strain increments and
damage evolution against the Python reference.}
    \label{fig:consrnn_vumat_validation_2}
\end{figure}

Before assessing this consistency, the deployment itself was validated at the
single integration-point level. Figures~\ref{fig:consrnn_vumat_validation}
and~\ref{fig:consrnn_vumat_validation_2} compare the embedded Fortran implementation
against the Python reference at Element~2852, Integration Point~2, and
Element~3089, Integration Point~3, respectively. In each case, the incremental
strain components $\Delta\varepsilon_1$ and $\Delta\varepsilon_2$ computed
inside the Fortran subroutine match the Python reference, obtained through
\texttt{numpy.diff} of the strain trajectory, to file-write precision (panels
(a) and (b)). The damage evolution predicted by the deployed Fortran ConsRNN
follows the Python reference implementation throughout the entire loading
history (panel (c)), with the pointwise difference between the two
implementations remaining below $6 \times 10^{-6}$ across all output steps,
100 and 80 respectively (panel (d)). These results confirm faithful deployment
of the trained ConsRNN model within the structural FEM context.

\begin{figure}[!h]
\centering
\includegraphics[scale = 0.85]{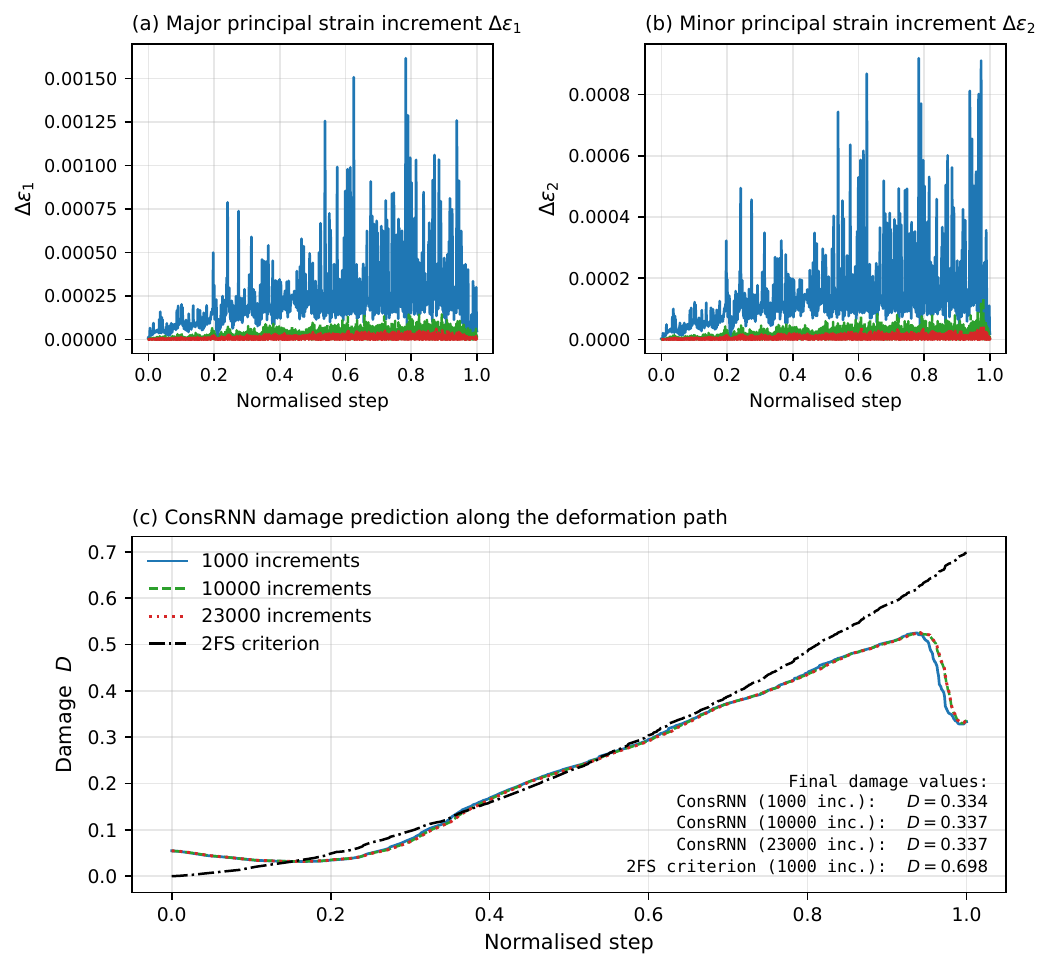}
\caption{ConsRNN damage predictions at Element~2852, Integration Point~2,
sampled at three temporal resolutions (1000, 10000, and 23000 increments):
(a)~major and (b)~minor principal strain increments; (c)~damage evolution
compared with the 2FS criterion.}\label{Simulation_Strain_Path_Element_2852_IP_2}
\end{figure}

\begin{figure}[!h]
\centering
\includegraphics[scale = 0.85]{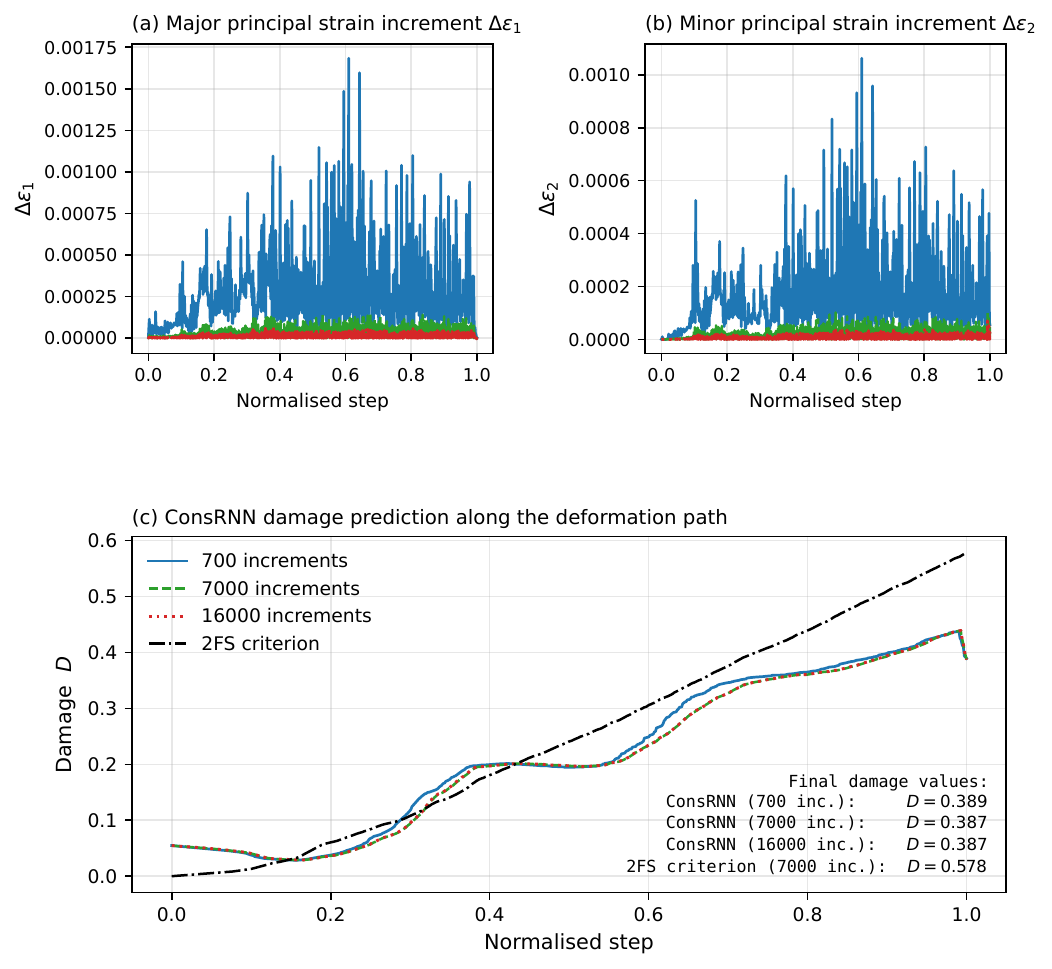}
\caption{ConsRNN damage predictions at Element~3089, Integration Point~3,
sampled at three temporal resolutions (700, 7000, and 16000 increments):
(a)~major and (b)~minor principal strain increments; (c)~damage evolution
compared with the 2FS criterion.}\label{Simulation_Strain_Path_Element_3089_IP_3}
\end{figure}

\begin{figure}[!h]
\centering
\includegraphics[scale = 0.85]{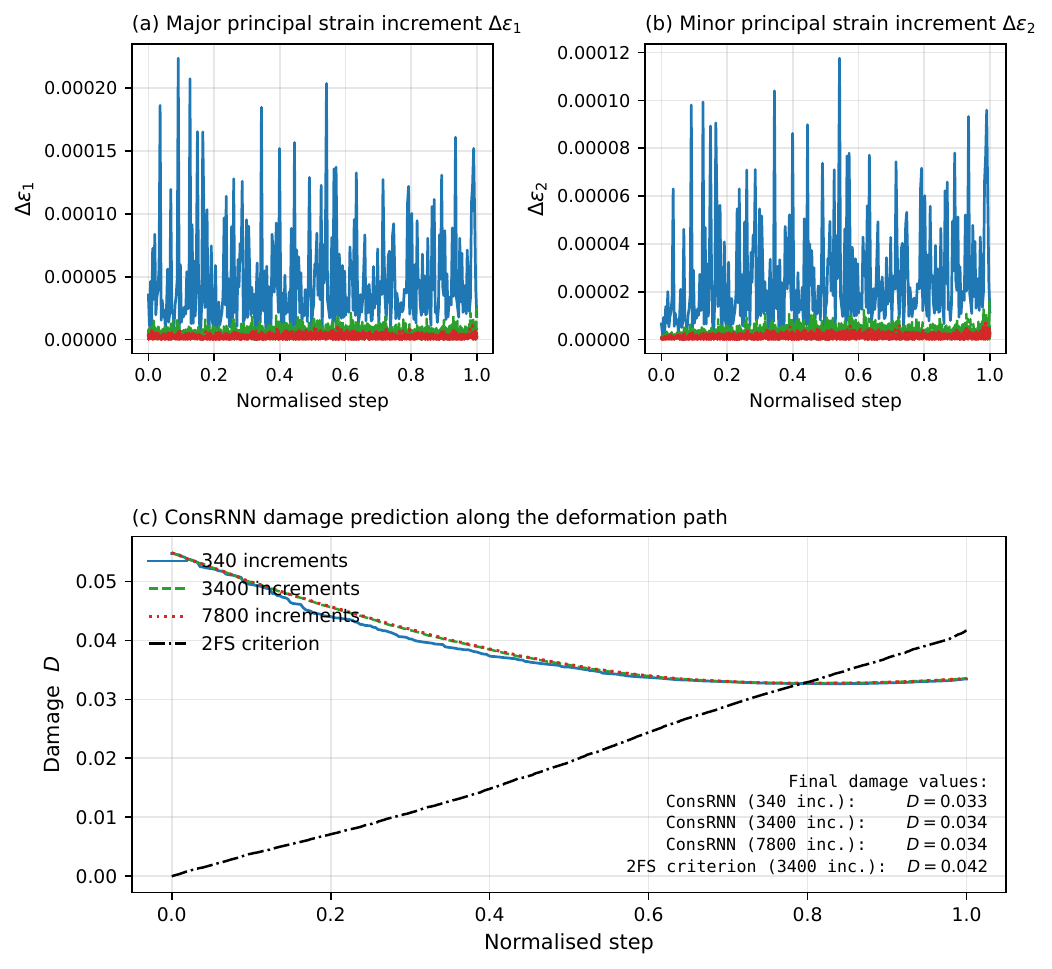}
\caption{ConsRNN damage predictions at Element~2750, Integration Point~1,
sampled at three temporal resolutions (340, 3400, and 7800 increments):
(a)~major and (b)~minor principal strain increments; (c)~damage evolution
compared with the 2FS criterion.}\label{Simulation_Strain_Path_Element_2750_IP_1}
\end{figure}

Building on this validated deployment, the embedded ConsRNN surrogate was
examined at three elements and integration points extracted from the structural
simulation, each with its own deformation history: Element~2852 at Integration
Point~2 (Fig.~\ref{Simulation_Strain_Path_Element_2852_IP_2}), Element~3089 at
Integration Point~3 (Fig.~\ref{Simulation_Strain_Path_Element_3089_IP_3}), and
Element~2750 at Integration Point~1
(Fig.~\ref{Simulation_Strain_Path_Element_2750_IP_1}). For each case, the same
VUMAT simulation was sampled at three temporal resolutions (1000, 10000, and
23000 increments; 700, 7000, and 16000 increments; and 340, 3400, and 7800
increments, respectively), producing strain-increment sequences with
substantially different time frequencies, as shown for the major and minor
principal strain increments in panels~(a) and~(b) of each figure. The ConsRNN
damage predictions were computed independently from these incremental principal
strain histories, and the traditional monotonically increasing 2FS fracture
criterion was extracted from the same simulation and overlaid along the
deformation path for comparison. The increment axis was normalised between
$0$ and $1$ to enable direct comparison across temporal resolutions and
deformation histories.

Despite the substantially different temporal resolutions, the damage
predictions in panel~(c) of each figure remain highly consistent along the
entire deformation path and converge toward similar final damage values:
$D = 0.334$--$0.337$ for Element~2852, $D = 0.387$--$0.389$ for Element~3089,
and $D = 0.033$--$0.034$ for Element~2750 across the respective increment
counts. The overlaid 2FS criterion accumulates damage monotonically by
construction and reaches different final values ($D = 0.698$, $0.578$, and
$0.042$, respectively), since the two criteria evolve damage on different
scales; the comparison is therefore qualitative rather than pointwise. Along
each deformation path, the ConsRNN predictions follow the same overall growth
trend as the 2FS reference, while exhibiting local non-monotonic segments that
reflect the history-dependent nature of the learned damage evolution. In
contrast to the 2FS curve, which is extracted at a single temporal resolution,
the ConsRNN trajectories remain nearly indistinguishable across resolutions,
which is precisely the property required for embedding within an explicit
solver. This behavior directly supports the discretisation-consistency
hypothesis of the proposed architecture under complex nonlinear deformation
histories.

One important point that should be emphasized is that the ConsRNN architecture
was trained only on bilinear loading paths, whereas the structural simulations
contain significantly more complex and nonlinear deformation histories. Despite
this distributional difference between training and testing conditions, the
embedded surrogate still produces stable and temporally consistent damage
predictions across varying increment counts. These observations demonstrate
that the proposed ConsRNN-based pipeline can be directly integrated into a
structural simulation environment and operate as an incremental surrogate
solver inside the simulation loop. Naturally, further improvements in
prediction accuracy are expected through the generation of more representative
nonlinear training paths and more specialized training datasets. Nevertheless,
the current results already demonstrate the feasibility of directly embedding a
recurrent neural-network-based surrogate model into a structural simulation
framework.

\begin{figure}[!htbp]
\centering
\includegraphics[scale = 0.90]{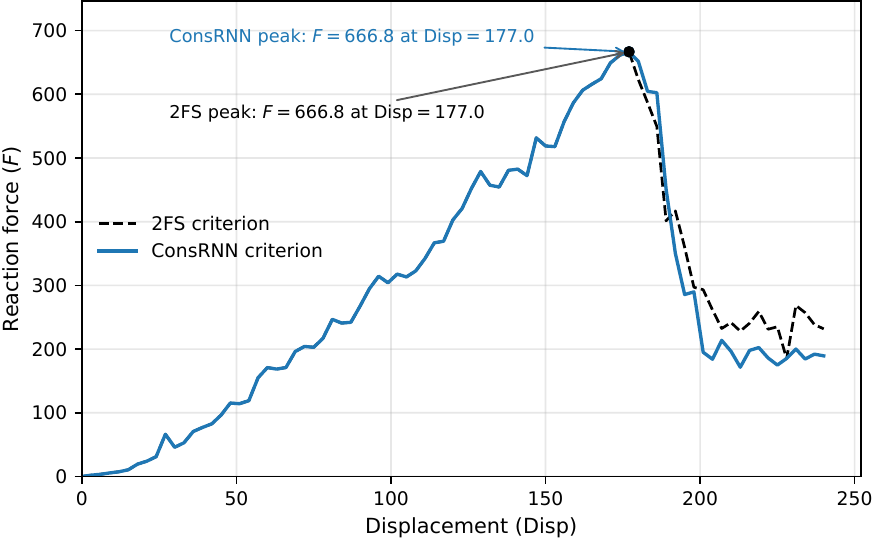}
\caption{Global force--displacement response with element erosion governed by 
the 2FS fracture criterion (dashed) and the embedded ConsRNN surrogate (solid). 
Both reach the same peak force ($F = 666.8$~kN) at the same displacement 
($\mathrm{Disp} = 177.0$~mm) and diverge only on the post-peak softening branch. 
The ConsRNN surrogate operates as a live element-deletion criterion inside the 
VUMAT.}
\label{F_D_Curve_Comparison}
\end{figure}

\subsection{Global Force and Displacement Curve}
\label{sec:Force-Displacement-comparison}
The global force--displacement response is the primary measure against which
fracture criteria in large-scale crash and collision analysis are validated,
since it is the quantity most reliably recorded in physical experiments and the
one that governs the predicted energy absorption of the structure. Its shape
encodes the essential structural behaviour: the pre-peak branch reflects the
combined elastic and plastic stiffness of the deforming members, the peak marks
the onset of fracture and the loss of load-carrying capacity, and the post-peak
softening branch captures the progressive failure and the residual resistance
as elements are eroded. A surrogate damage model is therefore only of practical
value if it can reproduce this response when embedded in a full structural
simulation, where each element experiences a non-proportional loading history
rather than a prescribed path. Recovering the force--displacement curve thus
serves as the decisive test of whether the model has learned a physically
meaningful damage evolution, and demonstrates that it can act as a drop-in
replacement for a traditional fracture criterion within the solver.

Figure~\ref{F_D_Curve_Comparison} compares the global force--displacement
response of the structure under two erosion criteria: the traditional 2FS
fracture criterion and the proposed ConsRNN surrogate, embedded directly
into the VUMAT as the element-deletion criterion. The two responses are
identical throughout the pre-peak regime: both simulations reach the same
peak force of $666.8$~kN at the same displacement of $177.0$~mm,
since the first element deletion occurs only beyond the peak, so that the
two criteria govern identical solution paths up to the onset of failure.
Both simulations accordingly exhibit the onset of load drop at the same
displacement, indicating that structural failure initiates at the same
stage of loading under either criterion. Beyond the peak, the curves
diverge moderately, with the ConsRNN run settling to a lower residual
load-carrying capacity than the 2FS run (a final force of $189.4$~kN
versus $231.8$~kN at the end of the prescribed loading), reflecting a
slightly more progressed element erosion on the post-peak softening
branch. This level of agreement is notable given that the ConsRNN model
was trained only on bilinear loading paths, yet here governs failure under
the non-linear, complex strain histories produced by the full structural
simulation. The result demonstrates that the proposed surrogate can be
embedded as a live fracture criterion inside a finite-element solver and
reproduce the global structural response of an established criterion,
while operating as an incremental damage predictor rather than an offline
post-processor.
\subsection{Discussion of the Results and Current Limitations}
\label{sec:results_discussion}

The consistency of the ConsRNN predictions can be observed at several levels.
For a single element under bilinear loading, Fig.~\ref{fig:path1} and Fig.~\ref{fig:path3} show that the
estimated damage remains consistent across different increment discretisations
of the same path. The same behaviour is seen at the single element, one
integration point level for the deformation-history based predictions obtained
from the structural simulation in Fig.~\ref{Simulation_Strain_Path_Element_2852_IP_2}, Fig.~\ref{Simulation_Strain_Path_Element_3089_IP_3}, and Fig.~\ref{Simulation_Strain_Path_Element_2750_IP_1}, where the proposed
transition-formula based architecture predicts consistent damage on non-linear
loading paths, although it was trained only on bilinear paths. Notably, even
though the incremental differences vary considerably along the non-linear paths
under different discretisations, the consistency in the estimates is preserved,
which is an important property for use within a structural simulation.
Additionally, Fig.~\ref{F_D_Curve_Comparison} shows that the proposed surrogate model produces a force
and displacement curve closely aligned with the traditional fracture criterion,
indicating that the model is a promising alternative. Taken together, these
results obtained inside the Abaqus/VUMAT environment support the central claim
of this paper.
Another behaviour that should be highlighted is that the predicted ConsRNN
damage values occasionally decrease during certain stages of the deformation
path before increasing again. This non-monotonic behaviour can be observed in
particular near the final stages of Fig.~\ref{Simulation_Strain_Path_Element_2852_IP_2} and Fig.~\ref{Simulation_Strain_Path_Element_3089_IP_3}.
Physically, damage evolution is generally expected to be irreversible, and the
predictions should therefore ideally evolve monotonically throughout the
loading history.
This behaviour originates directly from the current ConsRNN state-update
formulation given in Table \ref{tab:ConsRNN_vs_SimpleRNN}. In the present implementation, the
transition activation is computed using a hyperbolic tangent function, and is
therefore not constrained to remain larger than the previously accumulated
state. From the update rule, the incremental state evolution can be expressed as
\begin{equation}
\Delta S_t
=
S_t - S_{t-1}
=
(A_t - S_{t-1}) \, \|X_t\|_2 ,
\label{eq:state_increment}
\end{equation}
where $A_t$ is the transition activation and $\|X_t\|_2$ denotes the Euclidean
norm of the strain-increment vector. Since $\|X_t\|_2 \geq 0$, the sign of the
state increment is determined entirely by the term $(A_t - S_{t-1})$.
Consequently, whenever the transition activation becomes smaller than the
accumulated state, that is $A_t < S_{t-1}$, the state increment becomes negative
and the predicted damage temporarily decreases.
Enforcing physically compatible, monotonically increasing damage predictions
while preserving the consistency property of the transition formulation is a
non-trivial question that we identify as a direction for further research. The
two requirements are not independent: the same update mechanism that makes the
predictions insensitive to the discretisation of the strain path also allows the
state increment to take a negative sign when the transition activation falls
below the accumulated state. A constraint that simply forces the output to be
non-decreasing could therefore interfere with the consistency that is central to
the present formulation. A more principled treatment would instead seek a
transition formula that is monotonic by construction, for example through an
activation or accumulation rule that guarantees a non-negative state increment
without altering the path-consistent behaviour. We regard the design of such a
formulation, together with a study of its effect on the damage evolution and the
resulting structural response, as a worthwhile topic for future work.

\section{Conclusion}

In this study, we embedded a surrogate Consistent Recurrent Neural Network
(ConsRNN) directly within Abaqus finite element simulations through a Fortran
VUMAT subroutine. The model was developed and trained in Python and TensorFlow
on datasets extracted from finite element simulations, and the trained network
was then evaluated natively within the solver, with its weights, hidden state,
and incremental update computed in Fortran at every material point rather than
through an external coupling interface. The in-solver implementation reproduced
the reference Python predictions to within an agreement of order $10^{-6}$,
confirming that the embedding preserves the trained model rather than
approximating it. We further showed that ConsRNN operates as an incremental
solver, carrying the learned material evolution between strain increments and
producing a damage history that is consistent in time and largely independent
of the discretisation of the strain path.

The approach was validated on prescribed strain paths using single element
simulations and then assessed in full structural simulations of clamped shell
plates, where each element follows a non-proportional loading history that
differs in both shape and increment count. Although the network was trained
only on bilinear loading paths, it predicted element-level damage under these
more complex histories and reproduced a global force and displacement response
in close agreement with the traditional 2FS fracture criterion. This indicates
a degree of generalisation beyond the training regime, and demonstrates that a
learned damage model can be carried through a structural analysis without
re-evaluating a closed-form fracture locus at each step. Unlike a fixed
analytical criterion, the surrogate accumulates damage from the deformation
history itself, which offers a route to representing path-dependent behaviour
and to modelling materials or loading regimes for which no closed-form
criterion is readily available.

More broadly, this work demonstrates a complete and reproducible workflow for
developing a machine learning surrogate, training it on data from physical
simulations, and deploying it directly inside an established finite element
solver through a hard-coded user subroutine. We regard it as a representative
example of how a trained ML model can be tested and operated within engineering
simulation environments, and as a reference point for future efforts to
integrate learned constitutive and damage models into physics-based solvers.
Future work should extend the training data to richer loading histories and
quantify the model's accuracy and robustness across a wider range of structural
configurations and material systems.

\ack{The corresponding author wishes to express sincere gratitude to Mr. Valeh Farzaliyev for valuable discussions and insightful comments.}

\funding{Estonian Research Council, grant PSG754 (Coupled simulation model for ship crashworthiness assessment).}

\roles{Muhammed Adil Yatk{\i}n: conceptualization, methodology, software, validation, formal analysis, investigation, visualization, writing (original draft). Mihkel K{\~o}rgesaar: conceptualization, methodology, supervision, funding acquisition, writing (review and editing). Hasan Kurban: methodology, writing (review and editing).}

\data{The trained ConsRNN model weights, the standalone NumPy and Fortran implementations, the Abaqus/Explicit VUMAT deployment codes, and the evaluation scripts used to produce the figures in this paper are openly available on Zenodo at \url{https://doi.org/10.5281/zenodo.21207967}.}

\appendix

\section{Pseudocode of the damage-evolution networks}
\label{app:pseudocode}

Algorithms~\ref{alg:simplernn} and \ref{alg:consrnn} summarize the forward
passes of the baseline SimpleRNN and the proposed ConsRNN models as
executed at each material point. The full NumPy and Fortran reference
implementations, the trained model weights, and the Abaqus/Explicit VUMAT
deployment codes are openly available in the repository at \url{https://doi.org/10.5281/zenodo.21207967}.

\begin{algorithm}[H]
\caption{SimpleRNN damage evolution (baseline)}
\label{alg:simplernn}
\begin{algorithmic}[1]
\Require Trained weights $W_{x}, W_{h}, b_{h}, W_{d}, b_{d}$;
         accumulated strain path
         $\{(\varepsilon_{1,t},\, \varepsilon_{2,t})\}_{t=1}^{T}$
\Ensure Damage predictions $\{D_t\}_{t=1}^{T}$
\State $\mathbf{h}_0 \gets \mathbf{0}$ \Comment{initialize hidden state}
\For{$t = 1$ to $T$}
    \State $\Delta\varepsilon_{1,t} \gets \varepsilon_{1,t} - \varepsilon_{1,t-1}, \quad
            \Delta\varepsilon_{2,t} \gets \varepsilon_{2,t} - \varepsilon_{2,t-1}$ \Comment{strain increments}
    \State $\cot\phi_t \gets \Delta\varepsilon_{2,t} / \Delta\varepsilon_{1,t}$ \Comment{orientation of the loading direction}
    \State $\mathbf{x}_t \gets \left(\varepsilon_{1,t},\; \varepsilon_{2,t},\; \cot\phi_t\right)$
    \State $\mathbf{h}_t \gets \tanh\!\left(\mathbf{x}_t W_{x} + \mathbf{h}_{t-1} W_{h} + b_{h}\right)$ \Comment{full state overwrite}
    \State $p_t \gets \sigma\!\left(\mathbf{h}_t W_{d} + b_{d}\right)$
    \State $D_t \gets -\log\!\left(1 - p_t + \epsilon\right)$
\EndFor
\end{algorithmic}
\end{algorithm}

\begin{algorithm}[H]
\caption{ConsRNN damage evolution}
\label{alg:consrnn}
\begin{algorithmic}[1]
\Require Trained weights $W_{x}, W_{h}, b_{h}, W_{d}, b_{d}$;
         incremental strain path
         $\{(\Delta\varepsilon_{1,t},\, \Delta\varepsilon_{2,t})\}_{t=1}^{T}$
\Ensure Damage predictions $\{D_t\}_{t=1}^{T}$
\State $\mathbf{h}_0 \gets \mathbf{0}$ \Comment{initialize hidden state}
\For{$t = 1$ to $T$}
    \State $\Delta\boldsymbol{\varepsilon}_t \gets \left(\Delta\varepsilon_{1,t},\; \Delta\varepsilon_{2,t}\right)$
    \State $s_t \gets \operatorname{clip}\!\left(\lVert\Delta\boldsymbol{\varepsilon}_t\rVert_2,\; 10^{-15},\; 10^{2}\right)$
    \State $\hat{\mathbf{x}}_t \gets \Delta\boldsymbol{\varepsilon}_t / s_t$ \Comment{unit direction of the increment}
    \State $\tilde{\mathbf{h}}_t \gets \tanh\!\left(\hat{\mathbf{x}}_t W_{x} + \mathbf{h}_{t-1} W_{h} + b_{h}\right)$
    \State $\mathbf{h}_t \gets \mathbf{h}_{t-1} + \left(\tilde{\mathbf{h}}_t - \mathbf{h}_{t-1}\right) s_t$ \Comment{consistency update}
    \State $p_t \gets \sigma\!\left(\mathbf{h}_t W_{d} + b_{d}\right)$
    \State $D_t \gets -\log\!\left(1 - p_t + \epsilon\right)$
\EndFor
\end{algorithmic}
\end{algorithm}


\bibliographystyle{iopart-num}
\bibliography{Application}

\end{document}